\documentclass[sigconf]{acmart}

\usepackage{balance}
\usepackage{graphicx}
\usepackage{caption}
\usepackage[list=true]{subcaption}

\def\BibTeX{{\rm B\kern-.05em{\sc i\kern-.025em b}\kern-.08emT\kern-.1667em\lower.7ex\hbox{E}\kern-.125emX}}
    
\copyrightyear{2019} 
\acmYear{2019} 
\setcopyright{rightsretained} 
\acmConference[KDD '19]{The 25th ACM SIGKDD Conference on Knowledge Discovery and Data Mining}{August 4--8, 2019}{Anchorage, AK, USA}
\acmBooktitle{The 25th ACM SIGKDD Conference on Knowledge Discovery and Data Mining (KDD '19), August 4--8, 2019, Anchorage, AK, USA}
\acmDOI{10.1145/3292500.3330955}
\acmISBN{978-1-4503-6201-6/19/08}

\acmSubmissionID{rt0615p}

\begin{document}

\title{Assessing The Factual Accuracy of Generated Text}

%
\author{Ben Goodrich}
\authornote{Both authors contributed equally to this research.}
\email{bgoodrich@google.com}
\author{Vinay Rao}
\authornotemark[1]
\email{vinaysrao@google.com}

\author{Peter J. Liu}
\email{peterjliu@google.com}
\author{Mohammad Saleh}
\email{msaleh@google.com}
\affiliation{%
  \institution{Google Brain}
}

%
\renewcommand{\shortauthors}{Goodrich and Rao, et al.}

%
\begin{abstract}
We propose a model-based metric to estimate the factual accuracy of generated text that is complementary to typical scoring schemes like ROUGE (Recall-Oriented Understudy for Gisting Evaluation) and BLEU (Bilingual Evaluation Understudy). We introduce and release a new large-scale dataset based on Wikipedia and Wikidata to train relation classifiers and end-to-end fact extraction models. The end-to-end models are shown to be able to extract complete sets of facts from datasets with full pages of text. We then analyse multiple models that estimate factual accuracy on a Wikipedia text summarization task, and show their efficacy compared to ROUGE and other model-free variants by conducting a human evaluation study.
\end{abstract}


%
\maketitle

\section{Introduction}  \label{introduction}
Recently, there has been wide empirical success in text summarization \citep{rush2015neural, nallapati-2016, peterliu-msaleh-2017}, machine translation \citep{bahdanau-2014, google-nmt-2016, attention-is-all-you-need-2017}, dialogue response generation \citep{adversarial-dialog-gen-2017, milabot-2017, multiresolution-2016}, and other text generation tasks. For evaluation, these models generally rely on metrics like ROUGE (Recall-Oriented Understudy for Gisting Evaluation) \citep{rouge}, BLEU (Bilingual Evaluation Understudy) \citep{bleu} and perplexity \citep{perplexity} that measure locally constrained n-gram overlap. In this paper, we propose an automatic metric for evaluating the factual accuracy of generated text.
  
A fact \emph{f} is defined to be a relation tuple \emph{(subject, relation, object)}, where \emph{subject} has a binary \emph{relation} to \emph{object} and can be assumed to have been inferred from text or a knowledge base, e.g. \emph{Barack Hussein Obama II (born August 4, 1961) is an American politician who served as the 44th President of the United States from January 20, 2009 to January 20, 2017} implies a set of facts such as \emph{(Barack Obama, president of, United States), (Barack Obama, born on, August 4 1961)}.

In this paper, we limit our scope to the task of evaluating text summarization. To evaluate a text summarization model, we compare the ground-truth summary text, \emph{T} and the generated summary, \emph{G}. Let $f_t, f_g \in F$, and $F_T, F_G \subset F$ where $F$ is a set of relation tuples. 
\begin{align*}
F_T &= \{f_t \mid f_t \textit{ is inferred from ground-truth } T\}\\
F_G &= \{f_g \mid f_g \textit{ is inferred from generated-text } G\}
\end{align*}
The models used in the metric we propose do not make use of world knowledge (e.g. knowledge base) during inference, and to account for that we filter \(F_T\) and \(F_G\) by only considering claims made in \(G\) that can either be verified or refuted by statements in \(T\). Concretely, if $f_t = (subj_t, rel_t, obj_t) \in F_T$ and $f_g = (subj_g, rel_g, obj_g) \in F_G$
\begin{align*}
F_{T'} &= \{f_t \mid \exists f_g \text{ and } subj_t = subj_g, rel_t = rel_g \}\\
F_{G'} &= \{f_g \mid \exists f_t \text{ and } subj_g = subj_t, rel_g = rel_t \}
\end{align*}
We can then define factual accuracy \(fact_{acc}\) as the $precision$ between \(F_{T'}\) and \(F_{G'}\).
\begin{align}
  fact_{acc} &= \dfrac{|F_{T'} \cap F_{G'}|}{|F_{G'}|} \label{eq:fact_acc}
\end{align}
For example, consider ground-truth summary \emph{T}: \emph{Brad Pitt was born in 1963} and generated summary \emph{G}: \emph{Brad Pitt was born in 1961}. Then, $F_T$ = \emph{\{(Brad Pitt, born-in, 1963)\}}, $F_G$ = \emph{\{(Brad Pitt, born-in, 1961)\}}. The metric \(fact_{acc}\) = 0 indicates there is no factual consistency between the two summaries, whereas another metric like ROUGE-1 (1-gram overlap) measures 0.83. A real example is highlighted in Table \ref{tab:fact_inac_ex} where the summarization model commits such a mistake. It is important to be able to measure these mistakes accurately to aid in training factually accurate summarization models.
\begin{table*}[t]
\begin{tabular}{|p{2cm}|p{13cm}|}
\hline
    Target & Peter Duryea (July 14, 1939 -- March 24, 2013) was an American actor. He is best known for appearing in a pilot episode of Star Trek: The Original Series, ``The Cage'' (1964), most of which was reused in ``The Menagerie'' (1966), as Lieutenant Tyler. His father, Dan Duryea (1907 -- 1968), was also an actor.\\
\hline
    Output & Peter Duryea (April 23, 1907 -- March 24, 2013) was an American actor. He is best known for his role as Lt. Jose Tyler in the original Star Trek pilot, ``The Cage''\\
\hline
\end{tabular}
\caption{Example of factual inaccuracy noted in a summarization model \citep{peterliu-msaleh-2017}. In this example, the summarization model uses the subject (Peter Duryea)'s father, Dan Duryea's birthdate.}
\label{tab:fact_inac_ex}
\end{table*}
  
Extracting fact tuples from text has been previously studied in methods like OpenIE (Open Information Extraction) \citep{openie}. OpenIE extracts triplets with an unspecified schema, and the relation is usually the text linking the two entities. However, it does not leverage information from a knowledge base and leads to outputs that are hard to compare. For example, \emph{Person was born in that town} $\Rightarrow$ \emph{(Person, born in, town)}. But \emph{That town is the birthplace of Person} $\Rightarrow$ \emph{(Town, is the birthplace of, Person)}.
  
We standardize comparison by studying structured approaches to relation tuple extraction where the schema is fixed. We compare two approaches for fact extraction. One is a two-step process that first involves recognizing all the named entities in a sentence, and then classifying the relation for every pair of entities in the sentence \citep{sorokin, relation-extraction-selective-attention}. Our other approach is to use an end-to-end model with a Transformer-based architecture \citep{attention-is-all-you-need-2017} that is trained to output structured fact tuples. These models are described in Section \ref{model_metrics}. We create a new dataset for fact extraction using distant supervision \citep{distant-supervision} on Wikipedia text by cross-referencing facts from the Wikidata knowledge base \citep{wikidata}. To the best of our knowledge, this dataset is bigger and contains more relations and domains than previously used datasets for relation or fact tuple extraction.
  
Our main contributions are:
\begin{enumerate}
    \item We introduce model-based metrics to analyze the factual accuracy of generated text (Sec \ref{model_metrics}). We compare them against model-free metrics listed in Sec \ref{sec:model_free_metrics}.
    \item To train fact tuple extraction models, we release code (as part of the Tensor2Tensor\footnote{\url{https://github.com/tensorflow/tensor2tensor}} framework along with the model weights\footnote{\url{https://github.com/tensorflow/tensor2tensor/tree/master/tensor2tensor/data_generators/wikifact}}) and a large dataset (Sec \ref{dataset}) based on Wikidata and Wikipedia at \url{https://github.com/google-research-datasets/wikifact}.
      \item We show that a Transformer-based end-to-end fact extraction model is able to perform structured prediction of relation tuples, avoiding the need to split the process into multiple steps (named entity recognition, coreference resolution and relation classification). It is able to extract complete sets of facts from full pages of text in one pass.
    \item We conduct experiments to compare our proposed metric against human evaluation of factual accuracy of generated text (Sec \ref{human-evaluation}) and show that model-based metrics are better correlated with human judgment when compared to traditional metrics like ROUGE.
\end{enumerate}
Our models work under some limitations that are discussed in Sec \ref{sec:constraints}, and then Sec \ref{sec:future_work} discusses future work and ways to make our models more robust.
\section{Related Work and Motivation} \label{related-work}
Many evaluation metrics have been proposed for text generation tasks like BLEU \citep{bleu} and METEOR \citep{meteor} for machine translation and ROUGE \citep{rouge}, Basic Elements \citep{basic-elements} \& Pyramid \citep{pyramid} for text summarization. In \citet{evaluation-summary-2009}, the authors explain the different kinds of evaluation we can perform for summarization. They are broadly classified as extrinsic metrics that are specific to tasks (e.g. in summarizing a person, whether the date of birth has been included) and intrinsic metrics like grammaticality, coherency and non-redundancy that are based on the analysis of the summary. ROUGE, BLEU, sentence level F1 measures, etc are intrinsic content based metrics. \citet{texttruth} and other related works study ways to estimate the trustworthiness of answers to a question. With the recent shift towards using neural abstractive methods for text summarization and other text generation tasks, we believe that it is important to assess the factual accuracy of generated text. \citet{challenges_data_to_document} have also studied some extractive evaluative methods to assess the quality of generated text. This includes a Relation Generator, which predicts the relation between entities to assess the factual correctness of generated records. However, we introduce a much larger dataset and enable training end-to-end models that can extract fact triplets from text. We additionally perform detailed analysis of the fact extraction models.

Typical fact extraction pipelines are a multistage process consisting of part-of-speech tagging, named entity recognition \citep{ner-core-nlp, ner-neural, ner-bidi-lstm-cnn} that produces entities $\{e_i\}$ and then relation classification that predicts a relation $r_k$ for every pair of entities $(e_i, e_j)$. OpenIE \citep{openie} predicts a relation by linking the text connecting $e_i$ and $e_j$. Because it does not have a fixed schema, logical reasoning on its outputs are not possible. \citet{discovering-relations-between-nouns-categories} extend this to start with a fixed schema that can grow with more training, yet retain a consistent output surface form.

In this paper, we consider fact classification models with fixed schema. This idea has been studied in many previous works including \citet{multi-instance-multi-label-learning}, which considered datasets that have multiple relation labels for an entity pair, which each may have multiple instances in the input text. This was modeled as a graphical model over latent variables. \citet{relation-extraction-with-matrix-factorization} treated relation extraction as reasoning with matrix-factorization, and could work with surface-form texts and knowledge-base embeddings simultaneously. However, both of these works had datasets with very few types of relations, and were shown to work over limited domains. Recently, neural networks have been used for classifying relations. \citet{relation-extraction-selective-attention} used attention over multiple instances for the same entity pair to predict relations. \citet{sorokin} proposed to predict multiple relations in a sentence by using all the entity pairs and relation labels in the sentence as contextual input. We propose a simpler model where we classify relations between all the entity pairs in a sentence, without any additional context. We also make use of our proposed dataset that is bigger, more diverse and has more relation types. Our dataset also has article-level information that can be used to train models like in Section \ref{end-to-end}.
Since using two-step processes may be affected by compounding of errors across the models, some end-to-end approaches \citep{joint-entity-relation-table, joint-entity-relation-lstm} have been proposed, where the models extract entities and relations in one pass through the model. However, the method used in \citet{joint-entity-relation-table} required designing hand-crafted features and task-specific algorithms. \citet{joint-entity-relation-lstm} has a two-phase model that first extracts entity candidates and then predicts relations based on the parsed tree-structure of the sentence. We instead propose a sequence-to-sequence model that is able to output fact tuples directly, and does not require any feature engineering.

We found that the abstractive summarization models such as those described in \citet{peterliu-msaleh-2017} may generate sentences with factual inaccuracies (e.g. incorrect month in date of birth, wrong city in the state, etc.).  \citet{faithful-to-original} found that 30\% of summaries generated by a state-of-the-art summarization model contained factual inaccuracies. We found by running a large-scale experiment as described in Section \ref{human-evaluation}, that the summarization model had factual inaccuracy rate of approximately 17\%. We believe that this is because such mistakes are not heavily penalized by cross-entropy or n-gram based model losses and metrics.\\
As further motivation, we synthesized factually inaccurate samples by making simple corruptions to Wikipedia lead sections. We replaced mentions of dates (day and month only), locations or people with other entities of the same type in the text. For example, \textit{Barack was born on August 4, 1961 in Honolulu. He married Michelle on October 3, 1992 in Chicago.} becomes \textit{Barack was born on October 3, 1961 in Chicago. He married Michelle on August 4, 1992 in Honolulu.}. Table \ref{tab:models-on-synthetic-data} shows that model-free metrics such as ROUGE and OpenIE-based tuple comparison do not reflect the decline in factual accuracy due to such corruption as much as the model-based metrics do.
\begin{table}[h]
\begin{tabular}{|p{4cm}|p{2cm}|}
\hline
\textbf{Model} & \textbf{Accuracy}\\
\hline 
\hline
ROUGE-1 & 97.08\\
\hline
ROUGE-2 & 94.06\\
\hline
ROUGE-L & 96.02\\
\hline
OpenIE & 87.26\\
\hline
Binary Relation Classifier & 46.75\\
\hline
Relation Classifier & 59.30\\
\hline
E2E & 65.44\\
\hline
E2E-Reduced* & 57.10\\
\hline
\hline
Expected Accuracy** & 30.97\\
\hline
\end{tabular}
\caption{\footnotesize{Factual accuracy predicted by different metrics on synthesized samples. Binary Classifier is described in Sec \ref{sec:binary_relation_classifier}, Classifier in Sec \ref{classifier} and E2E is the end-to-end model described in Sec \ref{end-to-end}. E2E-Reduced* is a model where sentences where no entities are detected are filtered out from the input text. The Expected Accuracy** is calculated as the ratio of number of corrupted facts to the total number of facts in the article.}}
\label{tab:models-on-synthetic-data}
\end{table}

\section{Dataset} \label{dataset}
We create a dataset for fact extraction using distant supervision that is based entirely on the English Wikipedia corpus and the Wikidata knowledge base $W_{KB}$ \citep{wikidata}. Our distant supervisor is very similar to the one proposed by \citet{distant-supervision}. Although the inputs and labels for the classifier and end-to-end model are slightly different, we start by running an NER and co-reference resolution system$^{\ref{footnote-ner-coref}}$ on each Wikipedia article. The topic of that article is considered as the subject $e_s$. The other entities $e_j$ found in the article are considered objects. For every pair $(e_s, e_j)$, we say they are related if there is a relation $r_k$ such that the triplet $(e_s, r_k, e_j)$ is found in $W_{KB}$. We add this triplet to a set of positive examples $E_p$. If no such relation exists between $e_s$ and $e_j$, we add the triplet $(e_s, r_0, e_j)$ ($r_0$ denotes no-relation) to a set of negative examples $E_n$. 

\section{Model-based Metrics} \label{model_metrics}

In this section we describe models that can extract fact tuples from text and how we use them to define the factual accuracy metric as defined in Eq \ref{eq:fact_acc}. Given some input text $X$, we then extract claims made in $X$ as fact tuples.

\subsection{Named Entity Recognition (NER) + Relation Classifier} \label{classifier}
This approach consists of two steps, where we first recognize all the named entities $e_i$ from $X$ and then classify relations between entity pairs $(e_i, e_j)$.
\subsubsection{Named Entity Recognition}
Entities are real-world objects like people, locations, organizations etc that can be identified by a proper name\footnote{https://en.wikipedia.org/wiki/Named\_entity}. Entities can be identified with named-entity recognition (NER) systems like \citet{ner-bidi-lstm-cnn, ner-neural, ner-core-nlp} that take in $X$ and produce the set $\{e_i\}$. NER is followed by co-reference resolution\footnote{While we use an NER and co-reference resolution system that is not available to the public, the dataset we release (Section \ref{dataset}) has the positions of all the recognized and resolved entities that we use for training our classifier.\label{footnote-ner-coref}} \citep{coref-clark-manning, coref-marta, coref-heeyoung, coref-karthik}. Publicly available NER and co-reference systems include Stanford's CoreNLP\footnote{http://stanfordnlp.github.io/CoreNLP/coref.html} and NLTK\footnote{https://www.nltk.org/}.

\subsubsection{Relation Classifier}
For every pair $(e_i, e_j),\ e_i \not= e_j$ we consider all sentences $S_l$ in $X$ that contain both entities. The input to the classifier is then each of these sentences $S_l$. Because a sentence may contain multiple entities, we also add a prefix $SUBJ$ for $e_i$ and $OBJ$ to $e_j$ as a hint. For example, $X$ = \emph{Person1 was born in City1} becomes $S_l$ = \emph{SUBJ\{ Person1 \} was born in OBJ\{ City1 \}}. Unlike \citet{sorokin}, our classifier does not require additional context. Let $s_i$ be a token in the input sentence $S_l$ after NER, and $r^k$ denote the $k$th relation. Our classifier takes in input tokens $s_i$ that are first embedded onto a latent space, and then a stack of Transformer encoder-only layers process the whole sequence. A subsequent max-pooling layer selects one of these outputs that is then converted to a probability estimate of relations by a sigmoid operation. The exact series of operations can be viewed as:
\begin{align*}
    w_{1:n} &= embed(s_{1:n})\\
    h_{1:n} &= transformer\_encoder(w_{1:n})\\
    h_i &= max_i(h_{1:n}); h_i \in \Re^k\\
    p(r^k) &= \frac{1}{1 + e^{-h_i^k}} = sigmoid(h_i^k)
\end{align*}
Figure \ref{fig:classifier-model} also shows the architecture of this model. 

\subsubsection{Dataset preparation} \label{sec:classifier_data}
For every triplet $f$ in $E_p \cup E_n$, we have sentence(s)\footnote{There may be more than one sentence in the article that have mentions of the subject and object entity pair.} $S_l$ in the article that may describe the relation between $e_s$ and $e_j$. $S_l$ is processed so that subject and object are prefixed with ``\emph{SUBJ}'' and ``\emph{OBJ}'' as a hint to the model (Section \ref{classifier}). This leads to a dataset with 2.9 million positive examples and 34 million negative examples totaling to 45GiB on disk.
\subsubsection{$fact_{acc}$ with the Relation Classifier}
The classifier predicts a relation $r_k$ for each entity pair $(e_i, e_j)$. We extract such triplets from the ground-truth $T$ and generated text $G$, and use the definition from eq \ref{eq:fact_acc} to calculate the factual accuracy.

\subsection{End-to-End Extraction} \label{end-to-end}
\begin{figure}
\centering
\subcaptionbox{Classifier\label{fig:classifier-model}}{\includegraphics[width=0.40\textwidth]{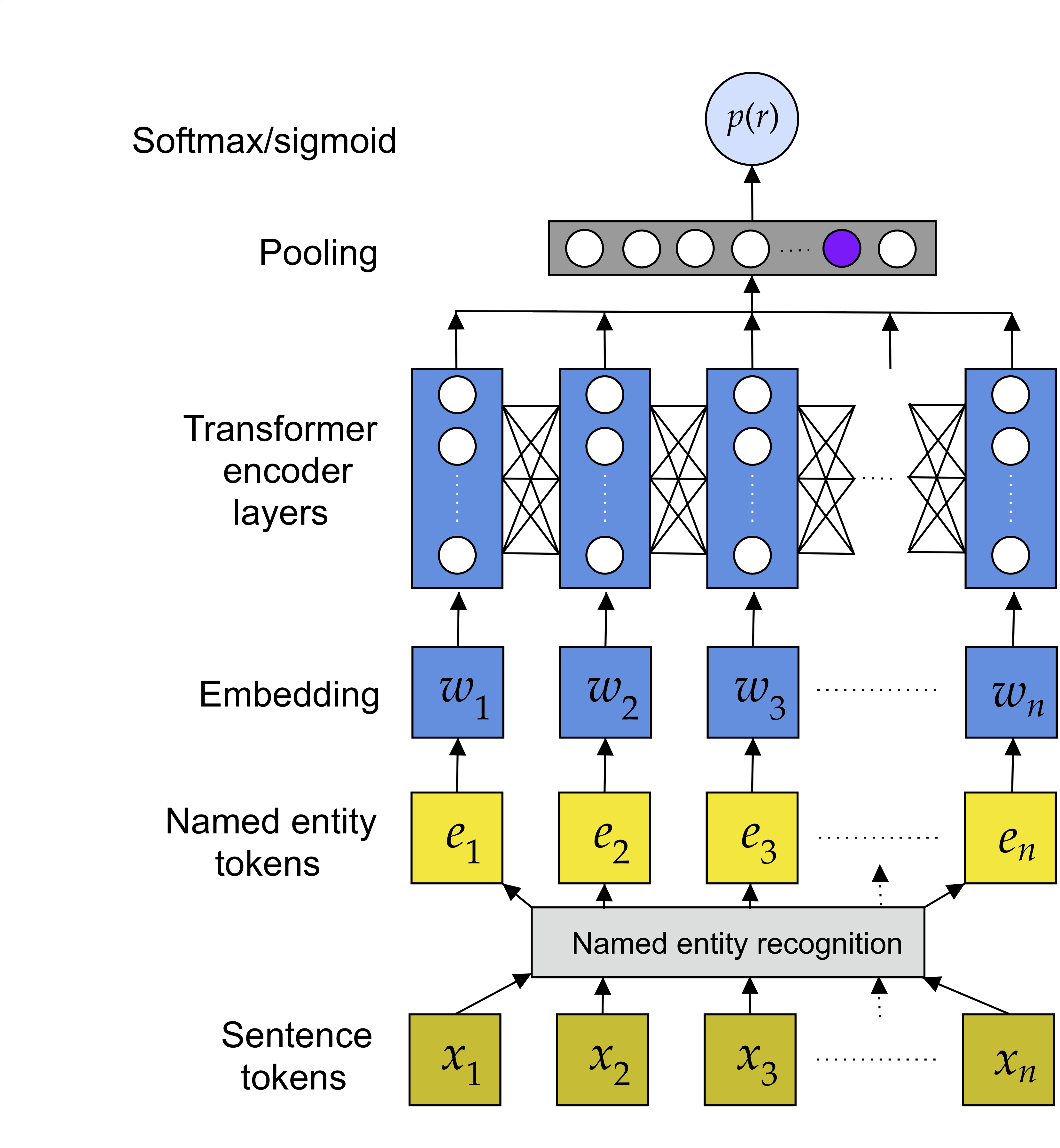}}%
\hfill
\subcaptionbox{Transformer encoder-decoder\label{fig:seq2seq-model}}{\includegraphics[width=0.40\textwidth]{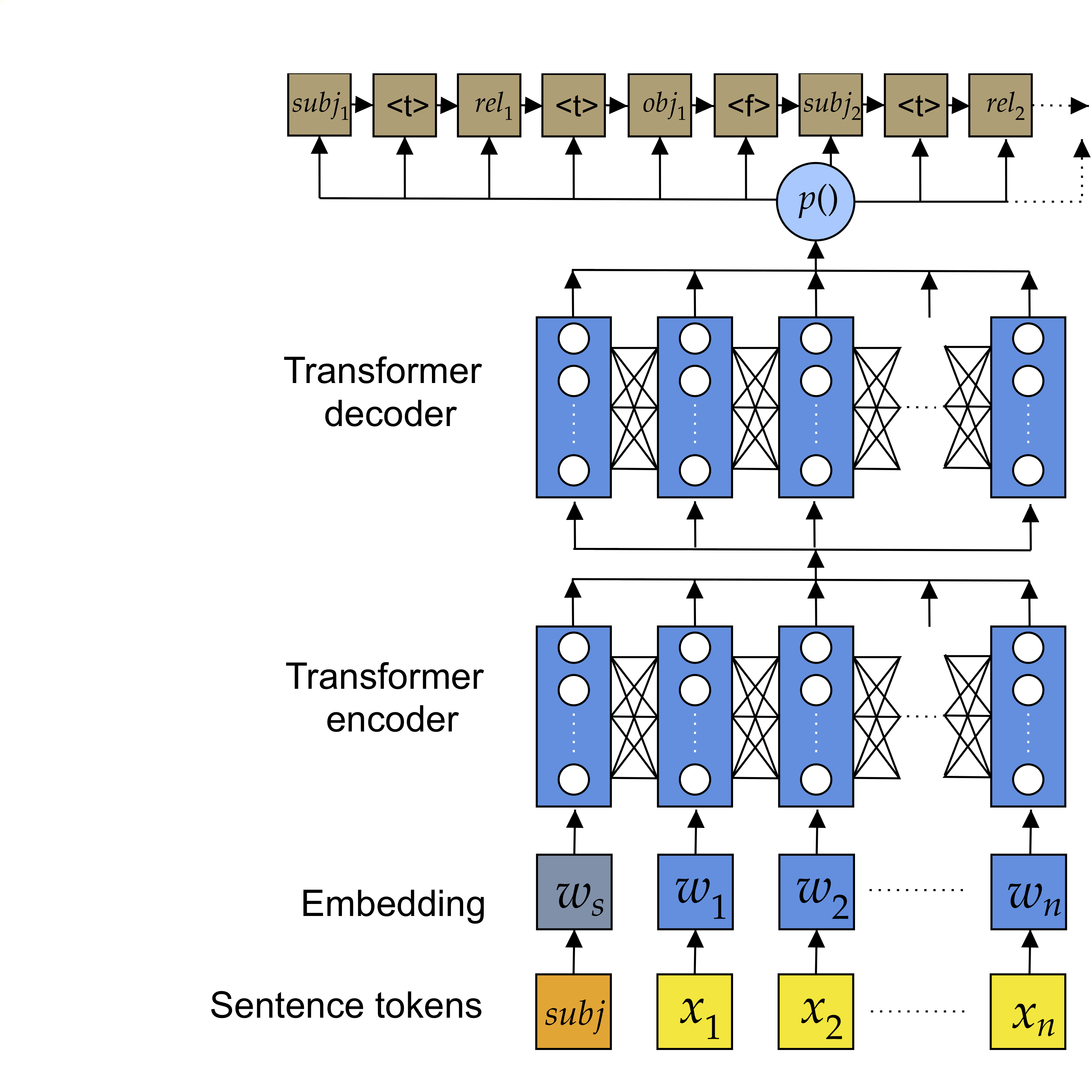}}%
\hfill
\caption{Fact extraction model architectures}
\label{fig:models}
\end{figure}

We propose an end-to-end fact extraction model to avoid compounding of errors across components in multi-stage approaches like Section \ref{classifier} \citep{Mccallum03anote}. This model also does not require any feature engineering or context. The input to the model is text $X$ of any length (sentence/paragraph/article) and the $subject$ entity $e_s$ prefixed to $X$. All the inputs tokens in $[e_s; X]$ are first embedded onto a latent space. A Transformer model consisting of a stack of encoder layers followed by decoder layers produces an output sequence of arbitrary length. A softmax operation is applied to every output token to define a distribution at every timestep. Figure \ref{fig:seq2seq-model} shows the architecture of this model. 
To encourage the model to have structured outputs, we train the model with labels that are a sequence of fact tuples. For example, if $X$ = `` \emph{Person1 was born in Country1. He was a painter}'', then the label, $Y$, for that input is ``\emph{Person1 $\langle$t$\rangle$ born in $\langle$t$\rangle$ Country1 $\langle$f$\rangle$ Person1 $\langle$t$\rangle$ profession $\langle$t$\rangle$ painter $\langle$end$\rangle$}'', where $\langle$t$\rangle$ separates tokens within the fact $f_i$ and $\langle$f$\rangle$ separates facts. For prediction, we perform a beam search over all the output timesteps, and continue decoding until $\langle$\emph{end}$\rangle$ is predicted. A length-penalty $\alpha$ controls the length of this prediction as in \citep{google-nmt-2016}.

\subsubsection{Dataset preparation}
If the input article text is $X$, every triplet $f_p$ in $E_p$ (we ignore the negative examples for end-to-end models because no relations between entity pairs is implied by no output by the model) is appended to the article's label $L$. $L$ will then contain a series of tokens that describe facts, with seperators between them. For example: $e_s \langle t \rangle r_1 \langle t \rangle e_1 \langle f \rangle e_s \langle t \rangle r_2 \langle t \rangle e_2 \langle f \rangle ...$ We also prepend the input text $X$ with $e_s$ ($[e_s; X]$) as a hint to the model for generating facts about $e_s$. This leads to a dataset with 2.5 million examples totaling to 1.5GiB on disk. \footnote{This dataset is made available at \url{https://github.com/tensorflow/tensor2tensor/tree/master/tensor2tensor/data_generators/wikifact}}
\subsubsection{$fact_{acc}$ with the End-to-End model}
The End-to-End model is able to produce a sequence of fact tuples in the form, \textit{$subj_1$ $\langle$t$\rangle$ $rel_1$ $\langle$t$\rangle$ $obj_1$ $\langle$f$\rangle$ $subj_1$ $\langle$t$\rangle$ $rel_2$ $\langle$t$\rangle$ $obj_2$}. It is trained to output relations from a fixed schema based on WikiData. Consider an output from this model, $\textit{Barack Obama} \langle t \rangle P69 \langle t \rangle Harvard$. $P69$ denotes `educated at'\footnote{https://www.wikidata.org/wiki/Property:P69}. These tuples are extracted from $T$ and $G$ to fit into the metric defined in eq \ref{eq:fact_acc}.

\subsection{NER + Binary Relation Classifier} \label{sec:binary_relation_classifier}
Similar to the typical relation classifier detailed in Sec \ref{classifier}, we define a classifier that predicts whether a pair of entities $(e_i, e_j)$ are related to each other through any relation. This allows for verifying that entities are related in both the ground-truth $T$ and generated text $G$, while being flexible enough to allow for any relation types. We also note that two entities can be related to each other in multiple ways. The inputs to this model are the same as Sec \ref{classifier}, but the model is expected to output $rel$ as
\begin{align*}
    rel = 
    \begin{cases}
            1 \colon e_i \text{ and } e_j \text{ are related}\\
            0 \colon \text{otherwise}
    \end{cases}
\end{align*}
\subsubsection{Dataset preparation}
Data for this model is generated with the same procedure detailed in Sec \ref{sec:classifier_data}. The only difference is the way we define the label $rel$. We consider entities $e_i$ and $e_j$ to be related if there is a relation $r_k$ such that $(e_i, r_k, e_j)$ is found in $W_{KB}$. 
\subsubsection{$fact_{acc}$ with the Binary Relation Classifier}
The model predicts $rel$ for each entity pair $(e_i, e_j)$, and we are able to extract a set of tuples of the form $(e_i, rel, e_j)$ from both $T$ and $G$. To use eq \ref{eq:fact_acc} to define the factual accuracy, we filter the set by considering only entity pairs $(e_i, e_j)$ that are found in both $T$ and $G$ to then compare the predicted label $rel$ between them.

\section{Model-free Metrics} \label{sec:model_free_metrics}
We describe model-free automatic metrics in this section. Unlike model-based metrics, they are not susceptible to changes in training data, and might be considered easier to interpret or understand.

\subsection{ROUGE} \label{sec:rouge}
ROUGE \cite{rouge} has been used as an automatic metric to judge the quality of generated text, and has shown to correlate well with human judgment of overall linguistic quality of the text.

\subsection{OpenIE} \label{sec:openie}
OpenIE \citep{openie} is a tool that can extract relation tuples from text, without a specified schema. We use it to extract sets of relation tuples from $T$ and $G$, and then compute the precision like in eq \ref{eq:fact_acc}.

\section{Model Experiments} \label{model-experiments}
In this section, we describe the methods we used to train and evaluate our relation extraction models. All of our proposed classifiers and end-to-end models have 6 Transformer layers and 1 embedding layer, with number of neurons (hidden layer size) set to 512. In the Transformer-based models, we use 8 attention heads. Our models are trained using the AdaFactor \citep{adafactor} optimizer. We use the publicly available Tensor2Tensor \citep{tensor2tensor}\footnote{https://github.com/tensorflow/tensor2tensor} framework for our experiments and will be releasing our code extensions as part of that framework. On our proposed dataset, the classifiers are trained for 50,000 iterations with batch-size of 1024 and the end-to-end models are trained for 50,000 iterations with batch-size of 256.\\
We evaluate classifiers and end-to-end models on our dataset. These results are presented in Table \ref{tab:models-on-dataset}. The end-to-end model is learning to recognize entities, resolving entity co-references, and reason about their relation in one pass through the model. To the best of our knowledge, we are not aware of other end-to-end structured relation extraction models and therefore do not include a comparison against other approaches. Some examples of extracting facts on our dataset are shown in \ref{appendix:fact-extraction}, where we include a comparison to OpenIE's triplet extraction.

\begin{table}[h]
\begin{tabular}{|p{3.3cm}|p{0.9cm}|p{0.9cm}|p{0.9cm}|}
\hline
\textbf{Model} & \textbf{P} & \textbf{R} & \textbf{F1}\\
\hline 
\hline
Binary Classifier* & 59.60 & 75.13 & 66.47\\
\hline
Relation Classifier & 63.49 & 68.64 & 65.96\\
\hline
E2E & 71.67 & 56.21 & 63.01\\
\hline
E2E-Reduced** & \textbf{72.16} & 61.03 & 66.13\\
\hline
\end{tabular}
\caption{\footnotesize{Performance (precision(P), recall(R), F1) of models on our proposed dataset grouped by classifiers and then end-to-end models. Binary Classifier is described in Sec \ref{sec:binary_relation_classifier}, Classifier in Sec \ref{classifier} and E2E is the end-to-end model described in Sec \ref{end-to-end}. The Binary Classifier* only considers the existence of a relation, and might not be directly comparable to the other models' performance. E2E-Reduced** is a model where sentences where no entities are detected are filtered out from the input text. The best model is marked in bold. We consider precision(P) as the measure that matches best with the definition of $fact_{acc}$}}
\label{tab:models-on-dataset}
\end{table}

We calculate precision and recall in the above experiments by matching ground-truth fact tuples exactly. This implies that the end-to-end model is not only learning to identify entities and resolve co-references, but also predict structured output, and its outputs can be used for reasoning. Their performance is competitive against relation classifiers while having a simple training and inference routine.

For each model, we sort and select the ten most frequent relation types that appear in our test sets. The $F1$ measure on these relations for classifiers are shown in Table \ref{tab:top-relations-classifier}, and end-to-end models are shown in Table \ref{tab:top-relations-end-to-end}.

\begin{table}[h]
\begin{tabular}{|p{3cm}|p{1cm}|p{1cm}|p{1cm}|}
\hline
\textbf{Relation} & \textbf{P} & \textbf{R} & \textbf{F1}\\
\hline
No relation & 0.9830 & 0.9817 & 0.9824\\
Country of citizenship & 0.6446 & 0.9394 & 0.7646\\
Date of birth & 0.9330 & 0.9850 & 0.9582\\
Country & 0.6049 & 0.9484 & 0.7386\\
Located in territory & 0.6260 & 0.8118 & 0.7069\\
Instance of & 0.5097 & 0.7015 & 0.5904\\
Place of birth & 0.6430 & 0.7436 & 0.6897\\
Member of sports team & 0.5179 & 0.9248 & 0.6640\\
Occupation & 0.5934 & 0.7770 & 0.6729\\
Date of death & 0.9163 & 0.9875 & 0.9506\\
\hline
\end{tabular}
\caption{\footnotesize{
    Precision (P), Recall (R) and F1 measure of the relation classifier (Section \ref{classifier}) on our test sets on ten most frequent relations. 
}}
\label{tab:top-relations-classifier}
\end{table}

\begin{table}[h]
\begin{tabular}{|p{3cm}|p{1cm}|p{1cm}|p{1cm}|}
\hline
\textbf{Relation} & \textbf{P} & \textbf{R} & \textbf{F1}\\
\hline
Country of citizenship & 0.8247 & 0.8359 & 0.8302\\
Instance of & 0.7212 & 0.6676 & 0.6934\\
Date of birth & 0.9342 & 0.9798 & 0.9564\\
Country & 0.8387 & 0.8267 & 0.8327\\
Cast member & 0.5889 & 0.4910 & 0.5355\\
Place of birth & 0.7012 & 0.7348 & 0.7176\\
Located in the administrative territorial entity & 0.7293 & 0.7700 & 0.7491\\
Member of sports team & 0.7045 & 0.7027 & 0.7036\\
Occupation & 0.5911 & 0.5774 & 0.5842\\
Educated at & 0.5432 & 0.7278 & 0.6221\\
\hline
\end{tabular}
\caption{\footnotesize{
    Precision (P), Recall (R) and F1 measure of our end-to-end model (Section \ref{end-to-end}) on our test sets on ten most frequent relations. 
}}
\label{tab:top-relations-end-to-end}
\end{table}

\section{Error Analysis of Model Predictions} \label{false-positive}
Distant supervision \citep{distant-supervision} is a way to create training data by using weak signals. In our dataset, we assign a relation label $r_k$ for every entity pair $(e_i, e_j)$ in the input text $X$ if the relation tuple $(e_i, r_k, e_j)$ exists in the Wikidata knowledge base $W_{KB}$. However, the sentence $S_l$ containing $(e_i, e_j)$ may not necessarily entail $r_k$. This leads to inaccurate estimates of the true-positive rate for our fact extraction models. We evaluate the effect of this distant supervision by gathering the set of facts extracted from our models that are marked false-positive by the distant supervision scheme. We present a pair of input text (Wikipedia articles) and facts extracted by our models to human evaluators, and ask them to mark a fact to be \emph{True} only if the relation tuple $(subject, relation, object)$ is implied by the input text. We asked two evaluators to score facts marked false-positive from a random set of 30 Wikipedia articles. We consider the fact to be true if both evaluators agree. We present the results in Table \ref{tab:false-positive}, where we can see the rate of false-positive facts that were marked true by the evaluators. This suggests that the end-to-end models could benefit by a better labeling scheme.

\begin{table}[h]
    \begin{tabular}{|p{2.5cm}|p{3cm}|}
    \hline
        \textbf{Model} & \textbf{\% True-positives} \\
    \hline
        End-to-end & \textbf{77.8}\\
        Relation Classifier & 46.6\\
    \hline
    \end{tabular}
    \caption{\footnotesize{Percentage of true facts that were inaccurately labelled wrong by the distant supervisor. The End-to-end model is the best model from Section \ref{end-to-end} and Classifier is the best from \ref{classifier}(Transformer-Sigmoid). The End-to-end model (in bold) predicts facts that are likelier to be true.}}
    \label{tab:false-positive}
\end{table}

\section{Evaluation of $fact_{acc}$ as a Metric} \label{evaluation}
In this section, we show the effectiveness of our proposed metric on judging the factual accuracy of generated text. We use the text summarization model proposed in \citep{peterliu-msaleh-2017} to generate lead sections of Wikipedia articles using the dataset and model in that paper, and compare the generated summary against the real lead section. In the following section, we describe the methodology used to compare human judgment of factual accuracy and how we compare our metric against that baseline.

\subsection{Human Evaluation} \label{human-evaluation}
\begin{center}
\begin{figure*}
  \centering
  \includegraphics[width=0.75\paperwidth]{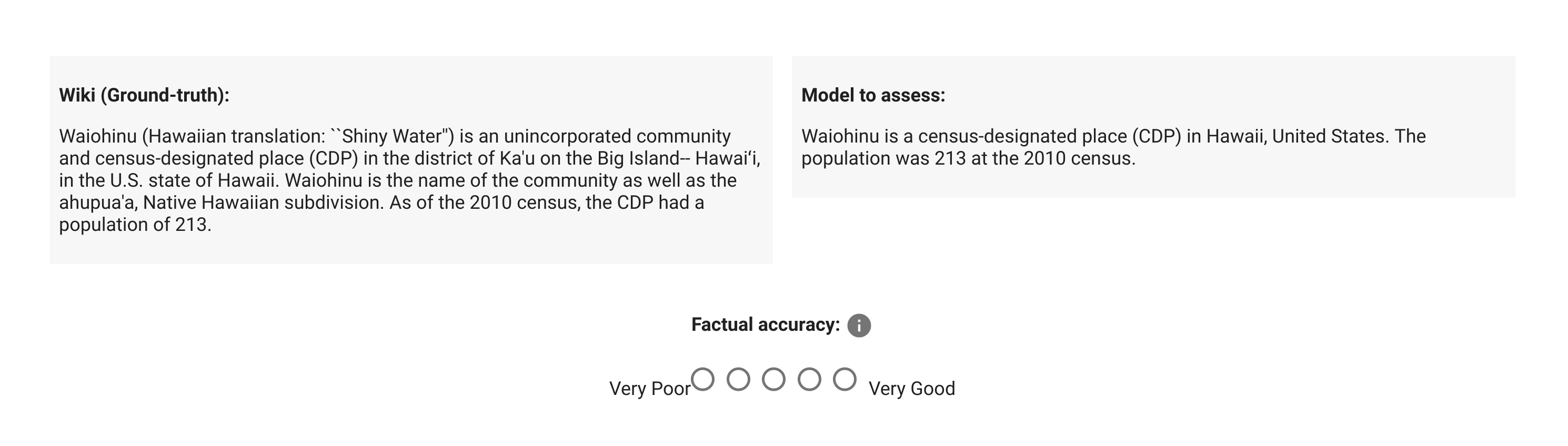}
  \caption{A screenshot of the interface presented to human evaluators to judge the factual accuracies of generated text. The ground-truth text is shown on the left, with the model generated text on the right. The evaluator is then asked to rate the factual accuracy of the generated text on a five point scale of `Very Poor' to `Very Good'}
  \label{image:fact_acc_human_eval}
\end{figure*}
\end{center}

Every claim made in the generated text $G$ can be considered to belong to one of three categories: supported by a sentence in ground-truth $T$, refuted by $T$ or cannot be verified by $T$. The evaluators were asked to only consider claims that are either supported or refuted by $T$. This ensures that no external knowledge is used in comparing $T$ and $G$, and ignores all claims that cannot be verified by $T$. Four evaluators were asked to rate 30 examples of generated text $G$ and then give it a score of 1-5 with 5 being highest factual accuracy. A special case is where the generated text has no verifiable claims. In this case, they were asked to give it a score of 1. Figure \ref{image:fact_acc_human_eval} shows the interface a human evaluator uses in our experiment.

We conduct the same experiment on two sets of data: first is a random sampling from summaries generated for Actors. We consider this an easier subset because we expect our fact extraction models to do well on this subset due to the summaries and Wikipedia lead sections generally containing relationships our models perform well on (see tables \ref{tab:top-relations-classifier} and \ref{tab:top-relations-end-to-end}). We present these results in Table \ref{tab:human-eval-actors}. We analyzed the inter-rater agreement on the scores given to each example, and found that Krippendorff's alpha (allows for ordinal rankings) was 0.6897. The second is a random sampling from all categories in Wikipedia. The results are presented in Table \ref{tab:human-eval-all}. The inter-rater agreement on this sample was found to be 0.7530.

We see that our end-to-end model (Section \ref{end-to-end}) has the best correlation on both subsets, indicating that it generalizes better to generated text. This may also be because the classifier suffers from a compounding of errors, where it is unable to predict relations if the NER system fails to recognize entities.

\begin{table}[h]
\begin{tabular}{|p{4cm}|p{2.8cm}|}
\hline
\textbf{Metric} & \textbf{Correlation with human scores}\\
\hline
\hline
ROUGE-1 & 0.583\\
\hline
ROUGE-2 & 0.639\\
\hline
ROUGE-L & 0.634\\
\hline
OpenIE & 0.258\\
\hline
$fact_{acc}$-Binary Classifier & 0.596\\
\hline
$fact_{acc}$-Relation Classifier & 0.523\\
\hline
$fact_{acc}$-E2E & 0.645\\
\hline
$fact_{acc}$-E2E-Reduced & \textbf{0.668}\\
\hline
\end{tabular}
\caption{\footnotesize{Spearman correlation of different metrics with human evaluation of factual accuracy on the `Actors' subset of summaries. ROUGE and OpenIE are described in Sec \ref{sec:model_free_metrics}, and the model-based $fact_{acc}$ metrics are described in Sec \ref{model_metrics}. The best metric is shown in bold.}}
\label{tab:human-eval-actors}
\end{table}

\begin{table}[h]
\begin{tabular}{|p{4cm}|p{2.8cm}|}
\hline
\textbf{Metric} & \textbf{Correlation with human scores}\\
\hline
\hline
ROUGE-1 & 0.384\\
\hline
ROUGE-2 & 0.435\\
\hline
ROUGE-L & 0.339\\
\hline
OpenIE & 0.128\\
\hline
$fact_{acc}$-Binary Classifier & 0.200\\
\hline
$fact_{acc}$-Relation Classifier & 0.250\\
\hline
$fact_{acc}$-E2E & 0.314\\
\hline
$fact_{acc}$-E2E-Reduced & \textbf{0.453}\\
\hline
\end{tabular}
\caption{\footnotesize{Spearman correlation of different metrics with human evaluation of factual accuracy on a random subset of summaries. ROUGE and OpenIE are described in Sec \ref{sec:model_free_metrics}, and the model-based $fact_{acc}$ metrics are described in Sec \ref{model_metrics}. The best metric is shown in bold.}}
\label{tab:human-eval-all}
\end{table}

\section{Conclusion}
\subsection{Limitations}\label{sec:constraints}
The dataset we create only makes use of sentences found in Wikipedia, and facts found in WikiData. This means that our models are biased to sentences structured to the neutral tone set in Wikipedia, and towards popular types of facts expressed in WikiData such as date of birth, profession, etc. Other sources of text may have more complex structures and styles of writing that may make it hard for our models to adapt to easily. An simple example of this is negating a binary relationship with `not', and different ways of expressing the same idea such as `wife/husband' instead of `spouse'. WikiData is an incomplete knowledge base, and this also leads to many sentences that in reality imply a fact to be marked containing no facts. This is a very typical problem faced by any work using distant supervision, and is combated with methods like active learning \citep{deepactivelearningner}.\\
It should be noted that ROUGE and to the best of our knowledge, most other automatic metrics, are also susceptible to changes in linguistic style and structure. However, elaborate labeling and bigger datasets will allow for our models to learn to overcome these challenges.

\subsection{Discussion and future work} \label{sec:future_work}
We have shown that our proposed metric is able to indicate the factual accuracy of generated text, and agrees with human judgment on our datasets. By leveraging a new dataset for both relation classification and end-to-end fact extraction, we also showed that classifiers and end-to-end models with straightforward architectures are able to perform competitive fact extraction.\\
Our end-to-end model avoids compounding of errors over sub-components typically used in other fact-extraction pipelines. We will release the code and datasets used to train this model, so that the proposed metric can be used to standardize comparison. We are in the process of building a bigger dataset that will contain multiple text domains, stronger human supervision and a larger collection of relation tuples that will help overcome many of the limitations discussed in the previous section (\ref{sec:constraints}). We encourage further development and use of this metric for automating the assessment of factual accuracy of generated text, and the development of better end-to-end models with structured outputs for fact extraction. 

\bibliographystyle{ACM-Reference-Format}
\bibliography{main}

\clearpage
\newpage
\appendix
\section{Appendix} \label{appendix}

\subsection{Reproducibility}
We release code to train our fact extraction models as part of the Tensor2Tensor framework\footnote{\url{https://github.com/tensorflow/tensor2tensor}} along with trained model weights
at \url{https://github.com/tensorflow/tensor2tensor/tree/master/tensor2tensor/data_generators/wikifact}.
A large fact extraction dataset (Sec \ref{dataset}) based on Wikidata and Wikipedia is made available \footnote{\url{https://github.com/tensorflow/tensor2tensor/tree/master/tensor2tensor/data_generators/wikifact}}.
To train our end-to-end and classifier models for fact extraction,
we use the hyper-parameter set ``transformer\_base'' defined in the Tensor2Tensor framework\footnote{\url{https://github.com/tensorflow/tensor2tensor/blob/master/tensor2tensor/models/transformer.py}}.
We further release code to use our end-to-end models as a fact extractor and calculate the factual accuracy metric at \url{https://github.com/tensorflow/tensor2tensor/tree/master/tensor2tensor/data_generators/wikifact}.

\subsection{Fact extraction example} \label{appendix:fact-extraction}
We include an example of facts extracted from text using our models where we compare it against OpenIE's \citep{openie} triplet extraction in Table \ref{tab:vs-openie}. This example illustrates the advantage of using structured approaches to fact extraction. OpenIE yields many triplets that mostly cannot be used for reasoning.
\begin{table*}[h]
    \begin{tabular}{|p{2cm}|p{13cm}|}
    \hline
        Input & Christopher Simon (born 5 June 1963) is an Australian actor and producer. Born in Sydney, Australia. He produced the film Miss You Already directed by Catherine Hardwicke. Simon is also a producer of such films as The Sweeney (2012 film) directed by Nick Love, Pusher, I, Anna, Still Life, Me and Me Dad, Boogie Woogie, The Proposition, Beyond the Ocean, The Trouble with Men and Women. He also produced short films by Joe Wright such as The End and Nick Love's Love Story. Simon's various television acting roles include Eddie in The Long Firm, Pedro in Gimme Gimme Gimme, Michael Hassan in The Bill, Lee Andersen in Casualty, Abdel in Lovejoy Samir in Ultimate Force, Da Souza in Lynda La Plante's Supply and Demand, Nathan Morgan in Wire In The Blood and he appeared in Lenny Henry in Pieces. Film acting roles include Room To Rent, The Delivery and O Jerusalem. Simon has acted in such plays as 12 Angry Men and Taking Sides both directed by Harold Pinter in London's west end, The Kitchen directed by Stephen Daldry at the Royal Court, the Amnesty award winning one man show When The Bulbull Stopped Singing for which he was nominated for the Acting Excellence Award (Best Actor) at the Edinburgh Festival Fringe, which premiered at the Traverse theatre and toured to Iran, New York and Jordan. Other theatre roles include Welcome to Ramallah, which toured York and London, at the Arcola and the Theatre Royal York, The Present at the Royal Court and the Bush, and Poor Superman at the Hampstead and the Traverse. \\
    \hline
        Targets & (Christopher Simon, date of birth, June 5 1963),
(Christopher Simon, country of citizenship, Australian),
(Christopher Simon, place of birth, Sydney)\\
    \hline
        OpenIE & (Abdel, is in, Ultimate Force),
(Casualty, Abdel in, Ultimate Force),
(Nathan Morgan, is In, Blood),
(Lee Andersen, is in, Casualty),
(Da Souza, is in, Lynda La Plante 's Supply),
(Simon 's various television acting roles, include, Eddie),
(Simon, is producer of, films as Sweeney directed by Nick Love),
(Simon, is also producer of, such films as Sweeney),
(Simon 's television roles, include, Eddie in Firm),
(Simon, is producer of, such films),
(Simon 's various television roles, include, Eddie),
(Eddie, is in, Long Firm),
(Simon, is producer of, such films as Sweeney),
(Simon, is producer of, such films as Sweeney directed by Nick Love),
(Michael Hassan, is in, Bill),
(Bill, Andersen in, Casualty),
(You, Already directed by, Catherine Hardwicke),
(Simon, is producer of, films),
(Simon, has, various television acting roles),
(Simon 's television acting roles, include, Eddie),
(Abdel, is in, Lovejoy Samir),
(Simon 's television roles, include, Eddie in Long Firm),
(Simon 's television acting roles, include, Eddie in Firm),
(Simon, is producer of, films as Sweeney directed),
(Simon 's various television acting roles, include, Eddie in Firm),
(Simon, is also producer of, films as Sweeney),
(Simon 's various television acting roles, include, Eddie in Long Firm),
(Simon, is, producer),
(Rent, To Room is, Delivery),
(Simon 's television roles, include, Eddie),
(Simon, is also producer of, films as Sweeney directed),
(Lynda La Plante, in, Supply),
(Pedro, is in, Gim),
...\\
    \hline
        Seq2Seq & (Christopher Simon, date of birth, June 5 1963),
(Christopher Simon, country of citizenship, Australian),
(Christopher Simon, place of birth, Sydney),
(Christopher Simon, occupation, Actor)\\
    \hline
        Classifier & (Christopher Simon, date of birth, June 5 1963),
(Christopher Simon, country of citizenship, Australian)\\
    \hline
    \end{tabular}
    \caption{Comparison of fact tuples extracted from this example, using OpenIE, our end-to-end model (Section \ref{end-to-end}), and our classifier (Section \ref{classifier}). A triplet consists of $(subject, relation, object)$.}
    \label{tab:vs-openie}
\end{table*}

\end{document}